\documentclass[conference]{IEEEtran}
\usepackage[T1]{fontenc}
\usepackage{newunicodechar}
\newunicodechar{—}{\textemdash}
\newunicodechar{’}{\textquoteright}
\IEEEoverridecommandlockouts
\usepackage{cite}
\usepackage{amsmath,amssymb,amsfonts}
\usepackage{algorithmic}
\usepackage{bbding}

\usepackage{marvosym}
\usepackage{float}
\usepackage{dutchcal}
\usepackage{hyperref}
\usepackage{graphicx}
\usepackage{color}
\usepackage{makecell}
\usepackage{datetime}
\usepackage{cases}
\usepackage{textcomp}
\usepackage[table]{xcolor}
\usepackage{cleveref}
\usepackage{booktabs}
\usepackage{multirow}
\usepackage{caption}
\usepackage{subcaption}
\usepackage{array}
\usepackage[ruled,vlined,linesnumbered]{algorithm2e}
\usepackage{eso-pic}
\AddToShipoutPictureFG*{\AtPageLowerLeft{\put(36,12){\parbox[b]{540pt}{\fontsize{6}{7}\selectfont
Author-accepted manuscript, ICME 2026. Supplementary material follows the main paper.\\
\textcopyright\ 2026 IEEE. Personal use of this material is permitted. Permission from IEEE must be obtained for all other uses, in any current or future media, including reprinting/republishing this material for advertising or promotional purposes, creating new collective works, for resale or redistribution to servers or lists, or reuse of any copyrighted component of this work in other works.
}}}}
\def\BibTeX{{\rm B\kern-.05em{\sc i\kern-.025em b}\kern-.08em
    T\kern-.1667em\lower.7ex\hbox{E}\kern-.125emX}}
\begin{document}

\title{EIB-Net: Entropy-Guided Information Bottleneck for Generalizable AI-Generated Image Detection}

\author{
\IEEEauthorblockN{Zhida Zhang}
\IEEEauthorblockA{\textit{School of Artificial Intelligence, } \\
\textit{UCAS \& NLPR, CASIA} \\
Beijing, China \\
zhida.zhang@cripac.ia.ac.cn
}
\and
\IEEEauthorblockN{Xinlei Ma}
\IEEEauthorblockA{\textit{School of Advanced Interdisciplinary Sciences, } \\
\textit{UCAS \& NLPR, CASIA} \\
Beijing, China \\
xinlei.ma@nlpr.ia.ac.cn
}
\and
\IEEEauthorblockN{Jie Cao \textsuperscript{*}}
\IEEEauthorblockA{\textit{NLPR, Institute of Automation,} \\
\textit{Chinese Academy of Sciences}\\
Beijing, China \\
jie.cao@cripac.ia.ac.cn
}

\thanks{\textsuperscript{*} is the corresponding author. This work was supported by the Beijing Natural Science Foundation (Grant No. L257008 \& L252145) and National Natural Science Foundation of China (Grant No. 62576338 \& 62550062 \& 32341009).}
}

\maketitle

\begin{abstract}
The proliferation of photorealistic AI-generated images demands robust detection methods that generalize across diverse generative models. While existing approaches target manipulation-based forgeries with local artifacts, generation-based images (e.g., from diffusion models) lack such traces, posing a fundamental challenge. We observe that generative models prioritize global semantics at the expense of local texture fidelity, making low-texture regions key indicators of synthetic origin. To exploit this, we propose EIB-Net, an Entropy-guided Information Bottleneck Network. EIB-Net introduces a novel Image Entropy (IE) metric to automatically select the most informative (lowest-entropy) patch, then processes it with a Variational Information Bottleneck (VIB) to learn compact, generalizable features. Extensive experiments on DIFF, DiffusionForensics, and GenImage benchmarks demonstrate state-of-the-art performance: EIB-Net achieves 85.7\% accuracy using only 2\% of training data, outperforming full-image baselines by over 15\%, and maintains robust cross-generator generalization (83.5\% average accuracy on GenImage). Furthermore, our entropy-guided patch selection (EGPL) consistently enhances diverse backbones (CNNs and Transformers), proving its practical value for data-efficient detection.
\end{abstract}

\begin{IEEEkeywords}
AI-generated Image Detection, Patch-based Analysis, Information Bottleneck, Generalization
\end{IEEEkeywords}

\section{Introduction}
\label{sec:intro}

In recent years, the rapid advancement of generative models, notably Generative Adversarial Networks (GANs) \cite{GAN} and the latest diffusion models (e.g., Stable Diffusion \cite{SDv1}, Midjourney \cite{Midjourney}), has pushed synthetic image realism to unprecedented levels. These text-to-image systems now produce content often indistinguishable from real photographs. While beneficial for creative applications, this capability simultaneously raises severe concerns regarding the misuse of synthetic media for disinformation, underscoring the urgent need for robust image authenticity verification techniques.

Most detection frameworks were originally developed for manipulation-based forgeries (e.g., deepfakes \cite{ff++}), which typically leave clear, localized artifacts like boundary inconsistencies. However, detecting pure \textbf{generation-based images} is fundamentally more challenging. Unlike tampered images, generated images lack explicit traces; their subtle statistical flaws are often distributed across the image, making traditional localization-based methods ineffective.

\begin{figure*}[t]
    \centering
    \includegraphics[width=0.75\textwidth]{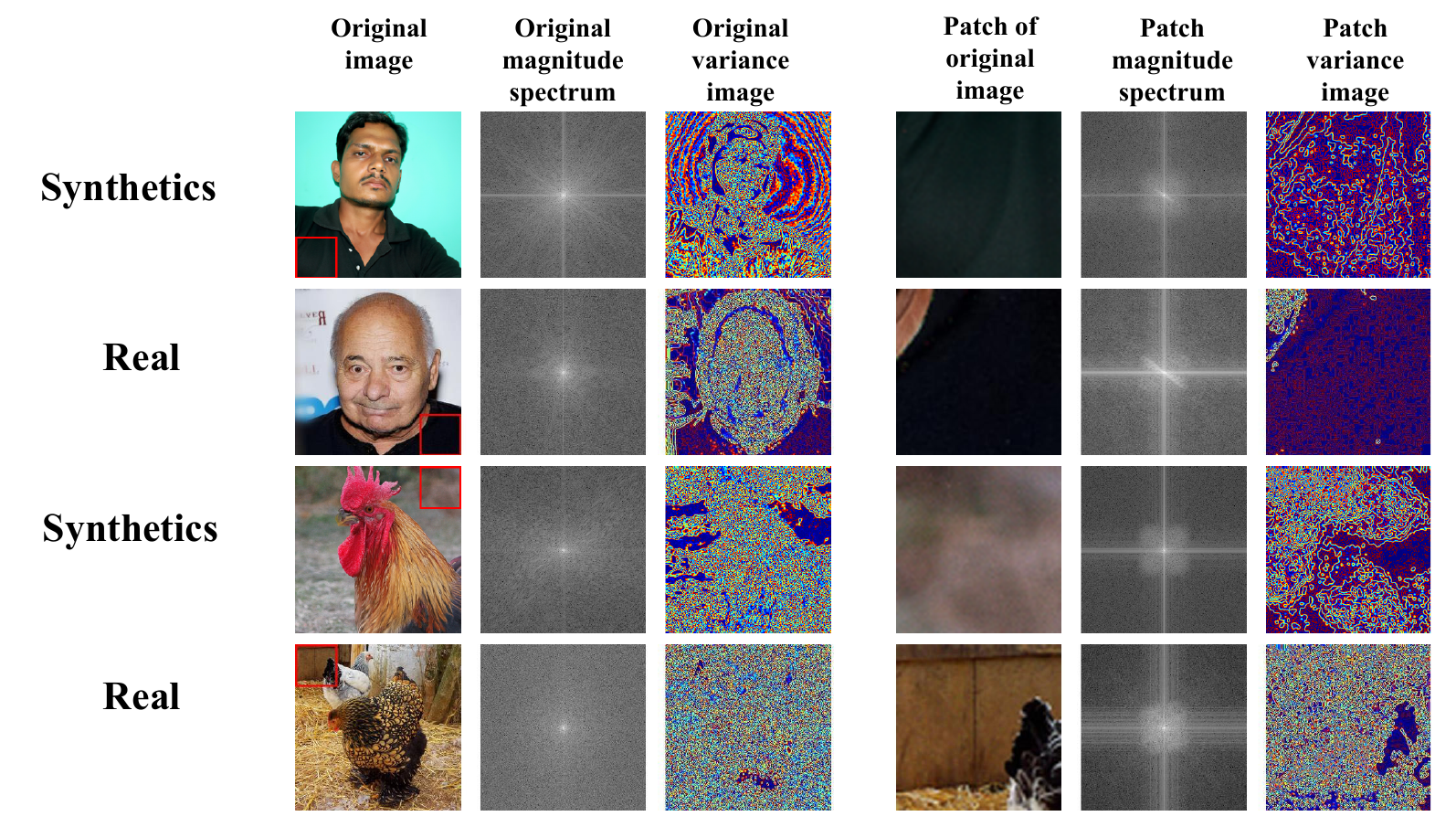} 
    \caption{Visual comparison of real and AI-generated images. The red box marks the lowest-entropy patch. The magnitude spectrum (middle) and variance image (right) show that real images contain richer high-frequency components and more complex local variance patterns.}
    \label{fig:patch_and_image}
    \vspace{-1em}
\end{figure*}

We identify the root cause of this failure: a mismatch in focus between generation and detection. Generative models are optimized to produce globally coherent semantics, often at the expense of local texture fidelity. In particular, \textbf{low-texture regions} (e.g., clear skies, smooth walls) receive weak supervision during training, leading to unnatural smoothness and a lack of characteristic high-frequency noise. This observation leads to our core insight: \textbf{the smoother, the faker}. Ironically, the most revealing artifacts of synthetic origin are found not in complex textures but in the simplest, most homogeneous areas.

This counter-intuitive insight is supported by the Gestalt principle of perception \cite{kohler1970gestalt}, which states that both humans and neural networks are biased toward global structure, often overlooking local details. By shifting the detection focus from global semantics to local statistical anomalies, we can build more robust and generalizable detectors.

To operationalize this insight, we propose the \textbf{E}ntropy-guided \textbf{I}nformation \textbf{B}ottleneck \textbf{Net}work (\textbf{EIB-Net}). EIB-Net implements a ``\textbf{select-then-compress}'' strategy through two novel components:
\begin{enumerate}
    \item \textbf{Entropy-Guided Patch Learning (EGPL)}: We design a novel Image Entropy metric to quantify local smoothness, automatically selecting the single most suspicious (lowest-entropy) patch as input.
    \item \textbf{Variational Information Bottleneck (VIB)}: This module compresses the patch representation, discarding irrelevant semantic content and forcing the model to learn compact, authenticity-relevant features.
\end{enumerate}
By focusing exclusively on the most artifact-prone region and suppressing content redundancy, EIB-Net learns to detect based on intrinsic texture fidelity rather than dataset-specific semantics.


Extensive experiments on DIFF, DiffusionForensics, and GenImage benchmarks validate EIB-Net's superiority. It achieves state-of-the-art accuracy in both intra- and cross-generator settings while exhibiting exceptional data efficiency—maintaining over 85\% accuracy with only 2\% of training data. Furthermore, the EGPL component proves model-agnostic, consistently enhancing diverse backbone architectures (CNNs and Transformers).

\noindent\textbf{Our main contributions are summarized as follows:}
\begin{itemize}
\item We propose \textbf{EIB-Net}, a novel patch-based detection framework combining \textbf{EGPL} and \textbf{VIB} for \textbf{robust, generalizable, and data-efficient} AI-generated image detection.
\item We introduce a new \textbf{Image Entropy (IE)} metric to guide patch selection toward smooth, artifact-prone regions, which significantly enhances convergence speed and generalization, especially under low-data regimes.
\item We demonstrate that the \textbf{EGPL} selection strategy is broadly applicable across diverse CNN and Transformer architectures, and that the full \textbf{EIB-Net} consistently achieves superior performance on challenging intra- and cross-generator benchmarks.
\end{itemize}

\section{Related Work}
\subsection{Generated Image Detection}
Early research on AI-generated image detection primarily focused on manipulation-based forgeries (e.g., deepfakes), typically employing specialized CNNs and attention mechanisms \cite{multi_task,cnn3,lisiam, duan2025dual}. However, these methods, which often rely on boundary or lighting artifacts, prove less effective against the highly photorealistic images generated by modern Latent Diffusion Models (LDMs), which lack explicit tampering traces.

To address the LDM challenge, several model-specific approaches emerged. Wang et al. \cite{DIRE} introduced Diffusion Reconstruction Error (DIRE), which measures discrepancies between an image and its reconstruction by a pre-trained diffusion model. Methods like AEROBLADE \cite{AEROBLADE} and LATENTTRACER \cite{LATENTTRACER} perform detection by analyzing LDM’s autoencoder representations without additional training. Other work leverages intrinsic statistical or structural flaws: Tan et al. \cite{NPR} propose Neighboring Pixel Relationships (NPR) to capture upsampling artifacts, while more recent generalizable approaches like \textbf{GenDet} \cite{zhu2023gendet} learn discriminative features from full images, which may still inadvertently encode semantic biases.

However, most approaches still rely on \textbf{model-specific reconstruction} (e.g., DIRE), target particular \textbf{geometric inconsistencies} \cite{Shadow}, or over-rely on \textbf{global semantic priors}. In stark contrast, our work is distinguished by its focus on \textbf{low-texture, low-entropy regions}. This deliberate departure from the established practice in camera-based forensic tasks (e.g., camera model identification) which\textit{high}-entropy (texture-rich) patches are typically selected to capture device-specific noise patterns. We posit that for \textit{synthesis} artifacts, the weakness of generative models lies in the \textit{under-supervised}, smooth areas—a novel hypothesis that guides our entropy-based patch selection.

\subsection{Mutual Information and Information Bottleneck}
The \textbf{Information Bottleneck (IB) principle} \cite{tishby2000information} seeks to compress input representations $X$ into a compact latent variable $Z$ while preserving task-relevant information about the target variable $Y$. This principle provides an effective means for enhancing generalization and robustness. Variational formulations of IB, such as the Variational Information Bottleneck (VIB) proposed in \cite{alemi2016deep,achille2018information}, adapt the IB principle to deep neural networks. These works reveal profound connections between IB, disentangled representation learning \cite{higgins2017beta}, and rate-distortion theory.

Recent research extends IB's utility across domains including reinforcement learning, adversarial robustness, and generative modeling \cite{versteeg2019bottleneck,sun2021ibood,wang2021ibadv}. In our work, the VIB module is not merely an add-on for regularization; it is integral to our strategy for overcoming semantic overfitting. By enforcing a compressed latent representation, the VIB actively \textbf{discards content-related information} from the selected image patch, compelling the network to retain only the minimal, task-relevant statistical cues for authenticity detection. This aligns with the recent direction of explicitly learning content-agnostic representations for media forensic tasks \cite{solopova2025}, and together with our low-entropy patch selection, forms a dual strategy operating at both the data and feature levels to achieve robust generalization.

\begin{figure*}[ht]
    \centering
    \includegraphics[width=1.0\textwidth]{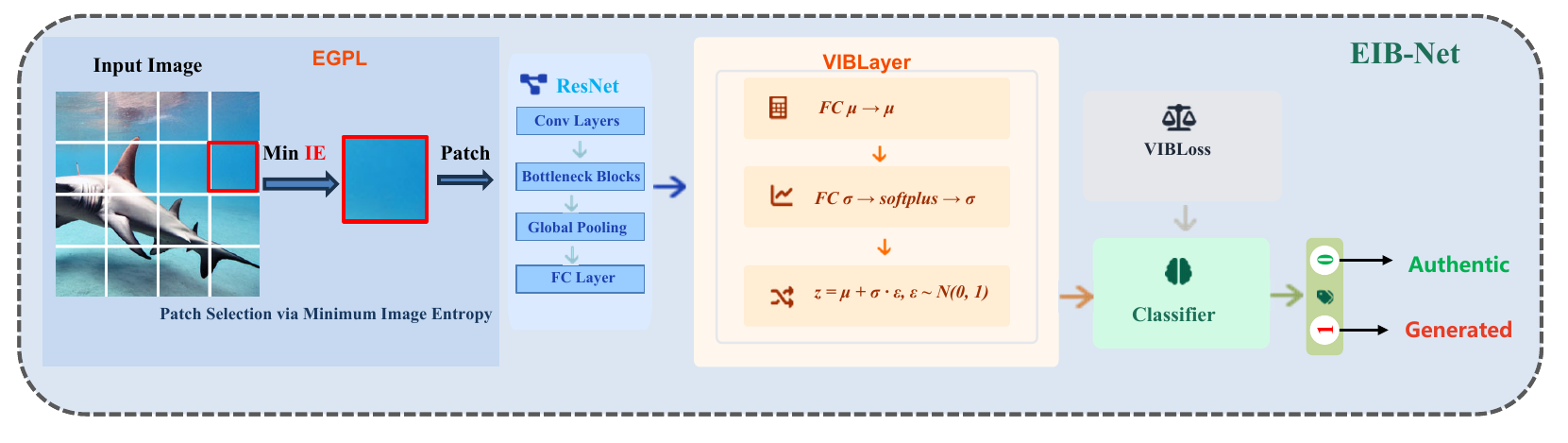}
    \caption{The pipeline of EIB-Net. Given an input image, the Entropy-Guided Patch Learning (EGPL) module computes the Image Entropy map and selects the patch $P_{\text{min}}$ with the minimum entropy. This patch is encoded by a ResNet backbone, and its features are compressed by the Variational Information Bottleneck (VIB) layer into a latent representation $\mathbf{z}$ for the final real/synthetic classification.}
    \label{fig:Pipeline}
    \vspace{-1em}
\end{figure*}

\begin{figure}[ht]
    \centering
    \includegraphics[width=0.45\textwidth]{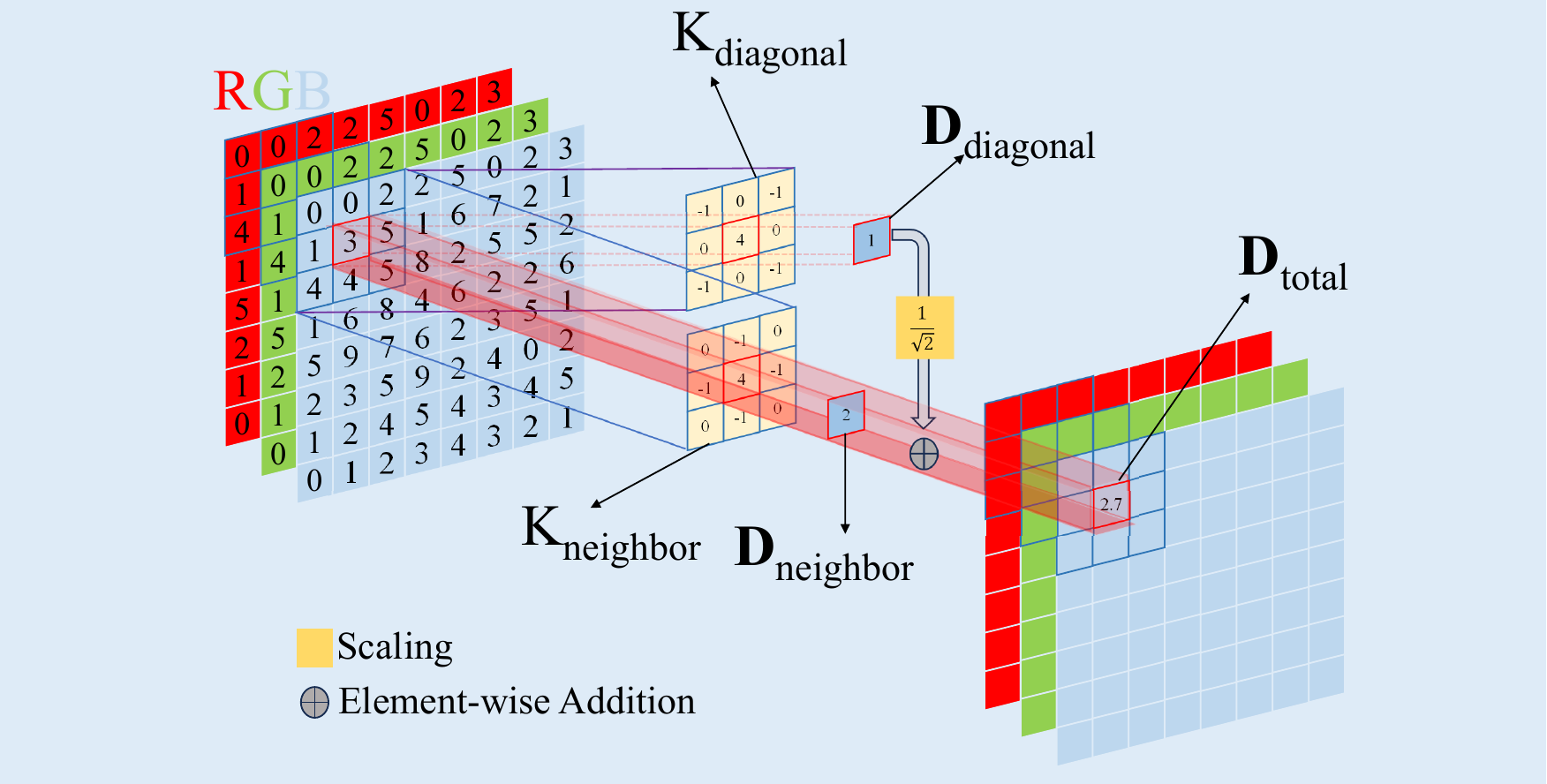}
    \caption{Illustration of Image Entropy (IE) computation. For each pixel in the patch, differences with neighboring and diagonal pixels are computed using $K_{\text{neighbor}}$ and $K_{\text{diagonal}}$. These differences are weighted, summed, and aggregated to form the total Image Entropy $\mathbf{D}_{\text{total}}$.}
    \label{fig:IE_Illustration}
    \vspace{-1em}
\end{figure}

\section{Method}

\subsection{Overview of EIB-Net}

Our proposed \textbf{EIB-Net} is designed to detect AI-generated images by directly countering the prevalent issue of \emph{semantic overfitting}. Instead of analyzing the full image—which often leads models to exploit dataset-specific content biases—EIB-Net forces the detector to base its decision on subtle, local statistical anomalies. As illustrated in \cref{fig:Pipeline}, this is achieved through a synergistic two-stage strategy:
\begin{itemize}
    \item \textbf{Entropy-Guided Patch Learning (EGPL)}: This front-end module implements our ``low-entropy hypothesis''. It computes a novel Image Entropy metric to automatically identify and select the single, smoothest image patch where generative artifacts are most likely to be exposed.
    \item \textbf{Feature Learning with Variational Bottleneck}: The selected patch is processed by a feature encoder (ResNet) followed by a \textbf{Variational Information Bottleneck (VIB)} layer. The VIB actively compresses the feature representation, discarding content-related information and preserving only the minimal statistical cues essential for authenticity classification.
\end{itemize}
This ``\textbf{select-then-compress}'' paradigm ensures the model learns from artifact-prone regions while being invariant to distracting global semantics, thereby achieving robust generalization.

\subsection{Entropy-Guided Patch Learning (EGPL)}

\subsubsection{Image Entropy Computation}

To locate the most vulnerable regions, we introduce an \textbf{Image Entropy (IE)} metric that quantifies local texture complexity across color channels. Given an RGB image $\mathbf{I} \in \mathbb{R}^{C \times H \times W}$ where $C=3$, we first divide it into $N$ non-overlapping patches $\{P_1, P_2, ..., P_N\}$.

The IE for each patch is computed by aggregating multi-directional intensity gradients within each color channel. We employ two Laplacian-style convolutional kernels:


\begin{equation}
    K_{\text{neighbor}} = 
    \begin{bmatrix}
        \begin{smallmatrix} 
        0 & -1 & 0 \\ -1 & 4 & -1 \\ 0 & -1 & 0 
        \end{smallmatrix}
    \end{bmatrix}, \quad
    K_{\text{diagonal}} = 
    \begin{bmatrix}
        \begin{smallmatrix} 
        -1 & 0 & -1 \\ 0 & 4 & 0 \\ -1 & 0 & -1 
        \end{smallmatrix}
    \end{bmatrix}.
\end{equation}
These kernels highlight intensity variations in adjacent and diagonal directions, respectively. The convolution is applied \textbf{independently to each of the three RGB channels}. For the entire image, the total variation map $\mathbf{D}_{\text{total}} \in \mathbb{R}^{C \times H \times W}$ is obtained by a weighted sum of the absolute responses:
\begin{equation}
    \mathbf{D}_{\text{total}} = |\mathbf{I} * K_{\text{neighbor}}| + \frac{1}{\sqrt{2}} |\mathbf{I} * K_{\text{diagonal}}|,
\end{equation}
where $*$ denotes the 2D convolution performed per channel.

The scalar entropy value $\text{IE}(P_i)$ for a patch $P_i$ is then defined as the sum of $\mathbf{D}_{\text{total}}$ over \textbf{all three color channels and all spatial locations} within that patch:
\begin{equation}
    \text{IE}(P_i) = \sum_{c=1}^{C} \sum_{(h,w) \in P_i} \mathbf{D}_{\text{total}}^{(c, h,w)}.
    \label{eq:patch_entropy}
\end{equation}
A lower IE value indicates a smoother, less textured region across color channels.

\subsubsection{Patch Selection}

Following our hypothesis that low-entropy areas are weak points for generators, the EGPL module selects the patch with the minimum entropy as the sole input to the subsequent detector:
\begin{equation}
    P_{\text{min}} = \arg\min_{P_i} \; \text{IE}(P_i).
    \label{eq:patch_selection}
\end{equation}
This operation focuses the model's attention precisely on the region where generative supervision is likely weakest and statistical anomalies are most pronounced

\subsection{Network Architecture with VIB}

The selected patch $P_{\text{min}}$ is resized to a standard resolution and fed into a ResNet-50 backbone (with the final fully-connected layer removed) to produce a feature vector $\mathbf{x} \in \mathbb{R}^d$.

To further purify this feature by stripping away residual semantic information, we employ a \textbf{Variational Information Bottleneck (VIB)} layer. The VIB models the distribution of a latent code $\mathbf{z}$ given $\mathbf{x}$. It first projects $\mathbf{x}$ into the parameters of a Gaussian distribution:

\begin{align}
    \boldsymbol{\mu} &= \mathbf{W}_\mu \mathbf{x} + \mathbf{b}_\mu, \\
    \log \boldsymbol{\sigma}^2 &= \mathbf{W}_\sigma \mathbf{x} + \mathbf{b}_\sigma - 5.
\end{align}
Here, the constant offset \(-5\) initializes \(\boldsymbol{\sigma}\) near zero, ensuring stable training in the early phases. The latent code \(\mathbf{z}\) is then sampled using the reparameterization trick: $\mathbf{z} = \boldsymbol{\mu} + \boldsymbol{\sigma} \odot \boldsymbol{\epsilon}$, where $\boldsymbol{\epsilon} \sim \mathcal{N}(0, \mathbf{I})$.
This stochastic bottleneck compels the model to form a compact representation \(\mathbf{z}\) that is maximally informative about the authenticity label while being minimally informative about the input \(\mathbf{x}\), thereby enforcing the information bottleneck principle.

\subsection{Loss Function}
The model is optimized end-to-end with a composite loss function $\mathcal{L}$ that balances classification accuracy, prediction consistency, and feature compression:
\begin{equation}
\mathcal{L} = \underbrace{\mathcal{L}_{\text{LS-CE}} + \alpha \cdot \mathcal{L}_{\text{JSD}}}_{\mathcal{L}_{\text{cls}}} + \beta \cdot \mathcal{L}_{\text{KL}}.
\label{eq:total_loss}
\end{equation}

\noindent\textbf{Classification Loss with Self-Regularization.} To promote generalization and calibrated predictions, the classification term $\mathcal{L}_{\text{cls}}$ combines a Label Smoothing Cross-Entropy loss ($\mathcal{L}_{\text{LS-CE}}$) with a Jensen-Shannon Divergence (JSD) regularization ($\mathcal{L}_{\text{JSD}}$) weighted by $\alpha=12$. $\mathcal{L}_{\text{LS-CE}}$ with smoothing factor $\epsilon=0.1$ prevents over-confidence. $\mathcal{L}_{\text{JSD}}$ encourages consistent output distributions and acts as an additional regularizer.
The overall classification loss is:
\begin{equation}
\mathcal{L}_{\text{cls}} = \mathcal{L}_{\text{LS-CE}} + \alpha \cdot \mathcal{L}_{\text{JSD}}.
\end{equation}
\noindent\textbf{Information Bottleneck Regularization.} The VIB module is regularized by the KL divergence between the learned latent distribution $q(\mathbf{z}|\mathbf{x})$ and a standard Gaussian prior $p(\mathbf{z})$:
\begin{equation}
\mathcal{L}_{\text{KL}} = \frac{1}{N} \sum_{i=1}^{N} \text{KL}\left[ q(\mathbf{z}_i|\mathbf{x}_i) \| p(\mathbf{z}_i) \right].
\label{eq:kl_loss_def}
\end{equation}

\noindent\textbf{Hyperparameters.} The trade-off between task performance and latent compression is controlled by $\beta$, set to $10^{-3}$ in all experiments.


\section{Experiments}

\subsection{Experimental Setup}
We utilize three benchmarks: \textbf{DIFF} \cite{DIFF} (faces), \textbf{DiffusionForensics} \cite{DIRE} (LSUN/ImageNet), and \textbf{GenImage} \cite{GenImage} (ImageNet-base). We evaluate on two settings: (1) \textbf{Intra-model:} Train/Test on the same generator distribution; (2) \textbf{Cross-model:} Train on one source (e.g., SDv1.4) and test on unseen generators. 
We report Accuracy (Acc) and provide full training details in the supplementary material.

\subsection{Intra-Model Generalization}
We evaluate in-domain robustness using a curated face-image subset, DIFF-Intra (see supplementary), varying training data ratios from 100

\textbf{Analysis.} The superiority of EIB-Net stems from its targeted learning strategy. While full-image models, especially ViT-B16, overfit to limited semantic cues, EGPL provides a strong \textit{inductive bias}. By focusing on low-entropy, artifact-prone regions, EGPL acts as an efficient data filter, enabling stable and sample-efficient learning from limited supervision.

\begin{table}[t]
    \centering
    \small
    \caption{Accuracy (\%) on DIFF-Intra under varying training data ratios. EIB-Net achieves superior performance across all data regimes, especially under low-data conditions. * indicates reproduced results.}

    \label{tab:DIFF}
    \setlength{\tabcolsep}{5pt} 

    \begin{tabular}{l|c|c|c|c}
        \toprule
        \textbf{Model} &  \textbf{100\%} & \textbf{20\%} & \textbf{5\%} & \textbf{2\%}\\
        \midrule
           ResNet50*& 98.66 & 90.86 & 86.69 & 82.67 \\

        \midrule
        RMT-S*& 98.81 & 93.27 & 84.66 & 70.66 \\

        \midrule
        ViT-B16* & 97.02 & 86.53 & 71.46 & 62.09\\

        \midrule
        Xception*& 98.85 & 97.09 & 76.28 & 70.12\\

        \midrule
        \cellcolor{blue!10} EIB-Net (Ours) & \cellcolor{blue!10} \bf 98.85 & \cellcolor{blue!10} \bf 97.17 & \cellcolor{blue!10} \bf 92.85 & \cellcolor{blue!10} \bf 85.69 \\
        \bottomrule
    \end{tabular}
\end{table}

\subsection{Cross-model generalization}
\subsubsection{DiffusionForensics}

Following \cite{DIRE}, models are trained on LSUN-Bedroom and tested on ImageNet samples from ADM and SDv1, a challenging cross-scene and cross-model setup. We report results at different training ratios to assess data efficiency (\cref{tab:DF_transposed}).

\textbf{Analysis.} EIB-Net maintains near-perfect validation accuracy ($>$99\% with 1\% data) and robust test accuracy ($>$80\% across both ADM and SDv1). In stark contrast, all baselines suffer a catastrophic drop in test performance, despite strong validation results. This confirms our hypothesis that baselines overfit to \textit{global, distribution-specific semantics} in the LSUN training set. EIB-Net, by leveraging \textit{local statistical anomalies} (via EGPL) and compressing semantics (via VIB), learns fundamentally more transferable features.

\begin{table}[t]
\centering
\small
\caption{Cross-model generalization on DiffusionForensics. Models are trained on LSUN-Bedroom and tested on ImageNet. Each row shows performance under a specific training ratio. Best results per row are in \textbf{bold}, EIB-Net column is shaded. }
\label{tab:DF_transposed}
\begin{tabular}{c|ccccc}
\toprule
\textbf{Train Ratio} & Res.50* & RMT-S* & ViT* & Xcep.* & \textbf{\cellcolor{blue!10}(Ours)} \\
\midrule
\rowcolor{blue!5} \multicolumn{6}{l}{\textit{Validation Accuracy (ADM\_VAL)}} \\
20\% & \textbf{100.00} & \textbf{100.00} & 96.10 & 92.65 & \cellcolor{blue!10}\textbf{100.00} \\
5\%  & 99.45 & 99.85 & 92.10 & 96.45 & \cellcolor{blue!10}\textbf{99.85} \\
1\%  & 91.80 & 98.75 & 84.15 & 90.55 & \cellcolor{blue!10}\textbf{99.00} \\
\midrule
\rowcolor{blue!5} \multicolumn{6}{l}{\textit{Test Accuracy (ADM)}} \\
20\% & 64.73 & 72.17 & 60.26 & 74.54 & \cellcolor{blue!10}\textbf{86.70} \\
5\%  & 71.29 & 81.20 & 57.31 & 76.55 & \cellcolor{blue!10}\textbf{87.50} \\
1\%  & 70.90 & 72.29 & 57.67 & 68.40 & \cellcolor{blue!10}\textbf{85.80} \\
\midrule
\rowcolor{blue!5} \multicolumn{6}{l}{\textit{Test Accuracy (SDv1)}} \\
20\% & 46.18 & 56.55 & 34.03 & 52.97 & \cellcolor{blue!10}\textbf{87.47} \\
5\%  & 43.27 & 57.67 & 34.91 & 52.88 & \cellcolor{blue!10}\textbf{87.87} \\
1\%  & 39.98 & 53.91 & 34.70 & 38.87 & \cellcolor{blue!10}\textbf{81.93} \\
\bottomrule
\end{tabular}
\vspace{-2em}
\end{table}

\subsubsection{GenImage}

We evaluate cross-generator generalization on the GenImage benchmark, training all methods exclusively on images synthesized by SD v1.4 and testing on samples from eight different generators.

As shown in \cref{tab:GenImage}, \textbf{EIB-Net} consistently outperforms both classical and state-of-the-art detectors across most generators. Notably, it achieves near-perfect accuracy on SD v1.4 (\textbf{99.89\%}) and the best average accuracy (\textbf{83.51\%}) across unseen generators.

\begin{table*}[t]
\caption{The cross-generator results on GenImage. All methods are trained only on the SD v1.4 subset. Accuracy (\%) is reported. The * symbol indicates the result of reproduction.}

\centering
\label{tab:GenImage}
\begin{tabular}{@{}l||c|cccccccc|c@{}}
\toprule
Method & Year & Mid & SDv1.4 & SDv1.5 & adms & glide & wukong & vqdm & BigGAN & Average\\
\midrule
ResNet18*&2016
& 62.33 & 99.75 &99.31 & 56.00 & 60.75 & 96.17 & 55.67 & 51.08 & 72.63\\
ResNet50* & 2016
& 61.25 & 99.92& 99.81& 57.83 & 69.58 & 98.67 & 59.67 & 53.75 & 75.06\\
Swin-T*  & 2021
&62.67&99.92&99.56&61.92&78.25&98.00&62.67&58.08&77.84\\
Xception*& 2017
&60.58&99.92&99.62&65.75&84.92&98.58&67.67&58.33&79.42\\
DIRE  & 2023
&60.20&99.90&99.80&50.90&55.00&99.20&50.10&50.20&70.66\\
GenDet & 2024
&89.60&96.10&96.10&58.00&78.40&92.80&66.50&75.00&81.56\\
PatchCraft  & 2024
&79.00&89.50&89.30&77.30&78.40&89.30&83.70&72.40&82.30\\

\cellcolor{blue!10}\textbf{EIB-Net} &\cellcolor{blue!10}\textbf{-}
&\cellcolor{blue!10}\textbf{76.92}&\cellcolor{blue!10}\textbf{99.89}&\cellcolor{blue!10}\textbf{99.89}&\cellcolor{blue!10}\textbf{72.17}&\cellcolor{blue!10}\textbf{80.25}&\cellcolor{blue!10}\textbf{99.88}&\cellcolor{blue!10}\textbf{64.25}&\cellcolor{blue!10}\textbf{74.83}&\cellcolor{blue!10}\textbf{83.51}\\
\bottomrule
\end{tabular}
\vspace{-1em}
\end{table*}

\subsection{Ablation Study}

We conduct ablation studies to validate the design of EIB-Net, with a focus on our core innovations: the Entropy-Guided Patch Learning (EGPL) module and the Variational Information Bottleneck (VIB).

\subsubsection{Analysis of EGPL Component}
\begin{figure}[ht]
    \centering
    \begin{minipage}[t]{0.49\textwidth}
        \centering
        \includegraphics[width=\textwidth]{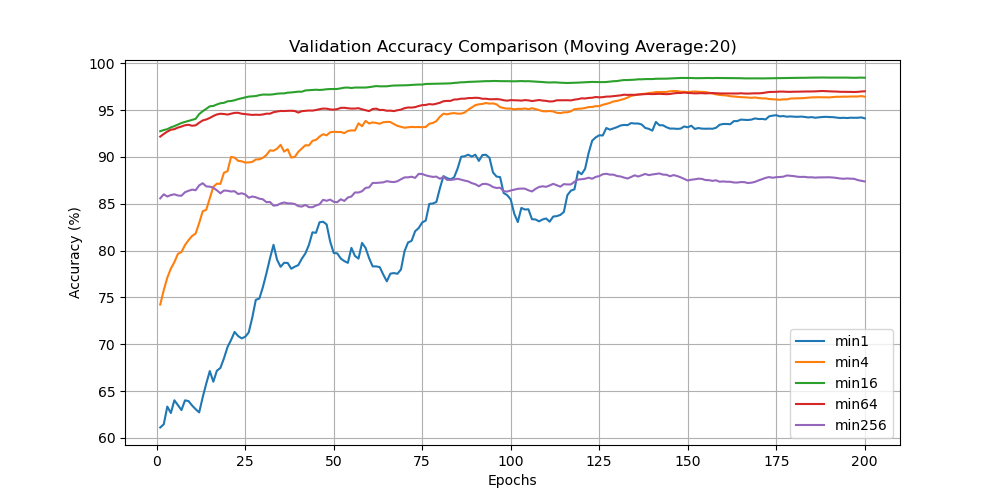}
        \caption*{(a) Accuracy vs. Patch Number}
        \label{fig:patchnum}
    \end{minipage}
    \hfill
    \begin{minipage}[t]{0.49\textwidth}
        \centering
        \includegraphics[width=\textwidth]{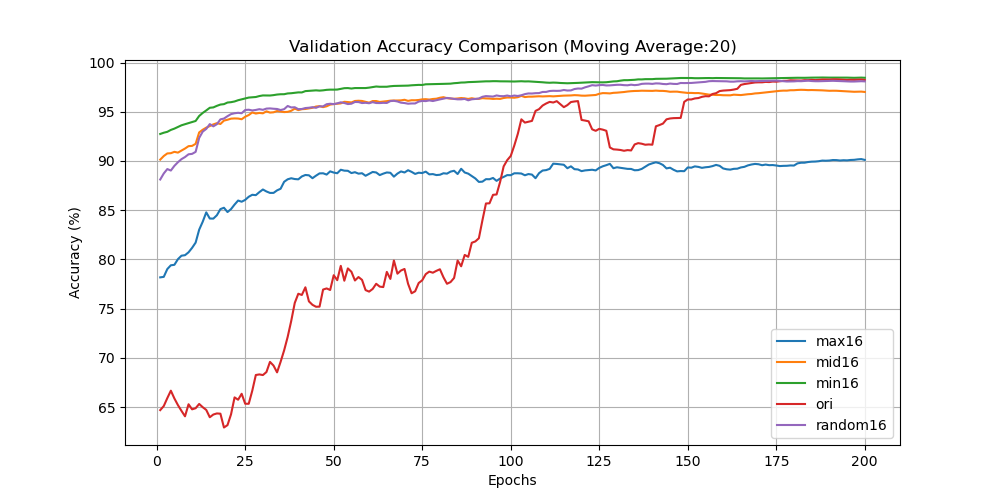}
        \caption*{(b) Accuracy vs. Patch Type}
        \label{fig:patchtype}
    \end{minipage}
    \caption{Validation accuracy of ResNet50 under varying patch configurations.
    (a) shows performance under different patch quantities, using the lowest-entropy patch from each split.
    (b) compares different patch selection strategies on a 16-patch split. Accuracy is smoothed with a moving average of 20 epochs.}
    \label{fig:patch_ablation}
    \vspace{-1em}
\end{figure}

\paragraph{Why Low-Entropy Patches?} We first ablate the patch selection strategy using a ResNet50 backbone. As shown in \cref{fig:patch_ablation}(b), selecting the \textbf{minimum entropy patch (min16)} achieves accuracy comparable to using the full image (ori) but with faster convergence. In contrast, selecting the \textbf{maximum entropy patch (max16)} results in the worst performance. This critical comparison provides direct empirical support for our ``low-entropy hypothesis'': high-entropy, texture-rich regions act as noisy distractors, while low-entropy, smooth regions are the most reliable indicators of generative artifacts.

\paragraph{How Many Patches?} We then analyze the granularity of patch splitting. Fig.~\ref{fig:patch_ablation}(a) shows that dividing the image into 16 patches (\textbf{min16}) offers the optimal trade-off. Too few patches (e.g., \textbf{min4}) may contain excessive irrelevant content, while too many (\textbf{min64}) may excessively fracture the artifact signal. The single full-image patch (\textbf{min1}) suffers from information loss due to rescaling.

\paragraph{Is EGPL Model-Agnostic?} To demonstrate the generality of our entropy guidance, we integrate EGPL into diverse backbone architectures. Table~\ref{tab:egpl_effect} shows that EGPL provides consistent and significant performance gains across CNNs (ResNet50) and Transformers (ViT-B16, RMT-S) on both intra- and cross-model benchmarks. This confirms that the benefit of focusing on low-entropy regions is a general principle, not an architecture-specific trick.

\begin{table}[t]
\centering
\caption{Impact of EGPL on different backbone architectures across datasets. DIFF(20\%) and DIFF(5\%) refer to intra-model validation accuracy with limited data. GenImage(Avg) is the average cross-model performance. EGPL improves generalization across all models.}
\label{tab:egpl_effect}
\setlength{\tabcolsep}{6pt}
\begin{tabular}{lccc}
\toprule
\textbf{Model} & \textbf{DIFF (20\%)} & \textbf{DIFF (5\%)} & \textbf{GenImage (Avg)} \\
\midrule
ResNet50 & 90.86 & 88.10 & 75.06 \\
\rowcolor{gray!10}
ResNet50 + EGPL & \textbf{95.94} & \textbf{92.00} & \textbf{80.62} \\
\midrule
RMT-S & 93.27 & 84.66 & 78.55 \\
\rowcolor{gray!10}
RMT-S + EGPL & \textbf{96.39} & \textbf{91.81} & \textbf{80.78} \\
\midrule
ViT-B16 & 86.53 & 71.46 & 77.13 \\
\rowcolor{gray!10}
ViT-B16 + EGPL & \textbf{92.16} & \textbf{85.92} & \textbf{80.46} \\
\midrule
EIB-Net (EGPL + VIB) & \textbf{97.17} & \textbf{92.85} & \textbf{83.51} \\
\bottomrule
\end{tabular}
\vspace{-1em}
\end{table}

\subsubsection{Analysis of VIB and Integrated Framework}
\paragraph{Role of the Information Bottleneck.} The VIB module is designed to compress features and discard semantic noise. 
Ablating the VIB module (i.e., using only EGPL) leads to a noticeable drop in cross-model generalization performance, validating its role in improving feature robustness.

\paragraph{Synergy of EGPL and VIB.} The final row of \cref{tab:egpl_effect} shows that the complete \textbf{EIB-Net (EGPL + VIB)} achieves the best overall performance, surpassing EGPL used alone on any single backbone. This demonstrates a clear synergy: \textbf{EGPL selects the most informative data (low-entropy patch), and VIB extracts the most informative features from it}, together forming a powerful ``select-then-compress'' pipeline that is highly efficient and robust.

\section{Conclusion}
\label{sec:conclusion}

This work tackles the critical issue of \emph{semantic overfitting} in AI-generated image detection. We propose a novel framework, \textbf{EIB-Net}, which introduces a ``\textbf{select-then-compress}'' strategy for robust and data-efficient detection. It first uses a novel \textbf{Image Entropy} metric to select the most artifact-prone (lowest-entropy) image patch, then applies a \textbf{Variational Information Bottleneck} to compress features and discard content semantics.

Extensive experiments show that EIB-Net achieves state-of-the-art cross-generator generalization while maintaining exceptional efficiency (e.g., $>$85\% accuracy with only 2\% training data). The entropy-guided selection is \textbf{model-agnostic}, consistently boosting diverse backbones.
Future work will extend this paradigm to video forgery detection and explore multi-patch aggregation for even broader robustness.

\clearpage
\begingroup
\setcounter{section}{0}
\setcounter{subsection}{0}
\setcounter{table}{0}
\setcounter{figure}{0}
\setcounter{equation}{0}
\renewcommand{\theHsection}{supp.\arabic{section}}
\renewcommand{\theHtable}{supp.\arabic{table}}
\renewcommand{\theHfigure}{supp.\arabic{figure}}
\renewcommand{\theHequation}{supp.\arabic{equation}}
\twocolumn[{\centering\LARGE Supplementary Material for \\
\textit{EIB-Net: Entropy-Guided Information Bottleneck for Generalizable AI-Generated Image Detection}\par\vspace{1.5em}}]


\section{Dataset Details}
\subsection{Overview of Evaluation Datasets}
We employ three public benchmarks and construct one customized subset for a comprehensive evaluation:
\begin{itemize}
    \item \textbf{DiffusionForensics} \cite{supp:DIRE}: Contains real images from LSUN-Bedroom \cite{supp:LSUN}, ImageNet, and CelebA-HQ \cite{supp:celebHQ}, alongside synthetic samples generated by pretrained diffusion models.
    \item \textbf{GenImage} \cite{supp:GenImage}: Includes over one million synthetic images generated by eight diffusion methods \cite{supp:nichol2022glide,supp:Wukong,supp:Midjourney,supp:LDMs,supp:VQDM,supp:classifierbootstrap}, with real images from ImageNet. The BigGAN subset is included in our cross-generator evaluation despite its non-diffusion origin.
    \item \textbf{DIFF} \cite{supp:DIFF}: A face-focused dataset with over 500K images synthesized by 13 diffusion models \cite{supp:podellsdxl,supp:freedom,supp:HPS,supp:hu2021lora,supp:ruiz2023dreambooth,supp:kim2022diffface,supp:kim2023dcface,supp:kawar2023imagic,supp:huang2023collaborative,supp:wu2023latent,supp:Midjourney}, under four generative conditions. Real faces are sourced from VoxCeleb2 and other celebrity datasets \cite{supp:DIFF_real1,supp:DIFF_real2,supp:DIFF_real3}.
\end{itemize}

\subsection{Construction of the DIFF-Intra Subset}
For the controlled intra-model generalization study in Sec. 4.2 of the main paper, we construct a balanced subset from the DIFF dataset, denoted as \textbf{DIFF-Intra}. It is constructed by randomly sampling an equal number of real and synthetic face images from the full DIFF dataset, ensuring a balanced and representative subset for intra-domain evaluation. This setup provides a clean and controlled distribution, specifically designed for evaluating in-domain generalization and data efficiency (few-shot learning) capabilities.

\subsection{Data Splits for Cross-Model Evaluation}
\begin{itemize}
    \item \textbf{DiffusionForensics} \cite{supp:DIRE}: We follow the standard split. Training/Validation: 80,000 real and 80,000 ADM-generated images from LSUN-Bedroom. Test: 10,000 ImageNet-based images each from ADM and SDv1 generators.
    \item \textbf{GenImage} \cite{supp:GenImage}: We use the official subset. Training is exclusively on the SD v1.4 subset (200k images). Evaluation covers all eight generators (including BigGAN), each with 20k images.
\end{itemize}

\section{Complete Experimental Configuration}
\subsection{Backbone Architectures and Baseline Training}
To fairly evaluate the general applicability of our EGPL module, we integrate it into four diverse, off-the-shelf backbone networks with minimal modification. The baseline training configurations for each are as follows:
\begin{itemize}
    \item \textbf{ResNet50} \cite{supp:resnet}: Initialized with ImageNet-pretrained weights. Trained with SGD (momentum=0.9, initial lr=1e-2, cosine annealing schedule). Uses JsdCrossEntropy loss (num\_splits=1, $\alpha=12$, label smoothing=0.1).
    \item \textbf{RMT-S} \cite{supp:rmt}: Uses the official implementation. Configured with AdamW (lr=1e-4, betas=(0.9,0.99), weight decay=1e-4) and exponential LR decay (rate=0.95 every 5 epochs). Uses BCEWithLogitsLoss.
    \item \textbf{ViT-B16} \cite{supp:vit}: Modified for binary classification. Trained with SGD (momentum=0.9, lr=1e-3, cosine annealing). Uses JsdCrossEntropy loss ($\alpha=12$).
    \item \textbf{Xception} \cite{supp:xception}: Optimized with AdamW (lr=1e-4) and exponential decay. Uses BCEWithLogitsLoss.
\end{itemize}
For all baselines, we only replace the final classification layer. This controlled setup ensures that any performance gain from EGPL is attributable to our method rather than extensive backbone-specific tuning.

\subsection{Configuration for EIB-Net (EGPL + VIB)}
The full EIB-Net framework uses ResNet50 as its feature encoder backbone, extended with our proposed modules. The specific configuration, which aligns exactly with the provided training code, is as follows:
\begin{itemize}
    \item \textbf{Optimizer}: SGD with momentum (0.9). No explicit weight decay is applied.
    \item \textbf{Learning Rate}: Initialized at $1 \times 10^{-2}$ (via \texttt{args.lr}), scheduled with a Cosine Annealing scheduler where \texttt{T\_max} equals the total number of epochs.
    \item \textbf{Batch Size}: 64 (via \texttt{args.BATCH\_SIZE}).
    \item \textbf{Training Epochs}: 200 (via \texttt{args.max\_epoch}).
    \item \textbf{Data Augmentation}: RandomResizedCrop(224), RandomHorizontalFlip(p=0.5), RandomErasing (p=0.6).
    \item \textbf{Loss Function}: $\mathcal{L} = \mathcal{L}_{\text{JSD}} + \beta \mathcal{L}_{\text{KL}}$, where $\mathcal{L}_{\text{JSD}}$ is the Jensen-Shannon Divergence enhanced cross-entropy with parameters $\alpha=12$, $\texttt{num\_splits}=1$, and label smoothing factor $\epsilon=0.1$. The VIB regularization weight is $\beta=10^{-3}$.
    \item \textbf{VIB Latent Dimension}: 256.
    \item \textbf{Reproducibility}: A fixed random seed (42) is used for all experiments.
\end{itemize}

\section{Algorithmic and Implementation Details}
\subsection{Training Pipeline}
\begin{algorithm}[ht]
\caption{Training Pipeline of EIB-Net}
\label{alg:eib_net}
\KwIn{Training image set $\mathcal{D}$, number of patches $N$, backbone network $f_\theta$, VIB layer $g_\phi$, classifier $h_\psi$}
\KwOut{Optimized parameters $\Theta = \{\theta, \phi, \psi\}$}
\textbf{Initialize} model parameters $\Theta$\;
\ForEach{epoch $ = 1 \to \text{MaxEpochs}$}{
    \ForEach{mini-batch $\{ \mathbf{I}_b, y_b \} \subset \mathcal{D}$}{
        \textbf{Step 1: Entropy-Guided Patch Selection (EGPL)} \;
        \ForEach{image $\mathbf{I}$ in $\mathbf{I}_b$}{
            Divide $\mathbf{I}$ into $N$ non-overlapping patches $\{P_1, ..., P_N\}$\;
            \ForEach{patch $P_i$}{
                Compute per-channel variation via Eq.~(1) in main paper\;
                Aggregate to get Image Entropy $\text{IE}(P_i)$ via Eq.~(2)\;
            }
            Select patch $P_{\text{min}} = \arg\min_i \text{IE}(P_i)$\;
            Resize $P_{\text{min}}$ to standard input size\;
        }
        Form selected-patch batch $\mathbf{P}_{\text{batch}}$\;
        
        \textbf{Step 2: Feature Extraction \& Compression} \;
        $\mathbf{x} \leftarrow f_\theta(\mathbf{P}_{\text{batch}})$ \tcp*{Backbone feature extraction}
        $\boldsymbol{\mu}, \boldsymbol{\sigma} \leftarrow g_\phi(\mathbf{x})$ \tcp*{VIB: Eq.~(3,4) in main paper}
        $\mathbf{z} \leftarrow \boldsymbol{\mu} + \boldsymbol{\sigma} \odot \epsilon, \; \epsilon \sim \mathcal{N}(0,I)$ \tcp*{Reparameterization}
        $\hat{\mathbf{y}} \leftarrow h_\psi(\mathbf{z})$ \tcp*{Classification}
        
        \textbf{Step 3: Optimization} \;
        Compute $\mathcal{L}_{\text{cls}}$ (JSD loss) and $\mathcal{L}_{\text{KL}}$ (KL divergence)\;
        $\mathcal{L}_{\text{total}} \leftarrow \mathcal{L}_{\text{cls}} + \beta \cdot \mathcal{L}_{\text{KL}}$ \tcp*{Eq.~(5) in main paper}
        Update $\Theta$ by descending $\nabla_{\Theta}\mathcal{L}_{\text{total}}$\;
    }
}
\end{algorithm}

\subsection{Implementation of Image Entropy}
The Image Entropy (IE) metric is computed efficiently on GPU to enable batch processing. Given an input batch of RGB images $\mathbf{I} \in \mathbb{R}^{B \times H \times W \times 3}$, the core steps are as follows:

\textbf{Convolution with Directional Kernels.} We employ two predefined $3\times3$ convolutional kernels, $K_{\text{neighbor}}$ and $K_{\text{diagonal}}$, which are mathematical Laplacian filters designed to highlight intensity variations in adjacent and diagonal directions, respectively (their explicit forms are given in Sec. 3.2 of the main paper). To process the RGB channels correctly, each kernel is repeated across the channel dimension (`repeat(3, 1, 1, 1)`). The 2D convolution is then applied \textbf{independently to each color channel} using PyTorch's \texttt{conv2d} function with the argument \texttt{groups=3}. This is equivalent to having three separate convolutional filters for the R, G, and B channels, ensuring that color information is processed independently without cross-channel mixing.

\textbf{Aggregation into Scalar Entropy.} The absolute responses of the two convolutions are summed to obtain a per-pixel, per-channel variation map $\mathbf{D}_{\text{total}}$, following Eq. (2) in the main paper. The final scalar entropy value $\text{IE}(P)$ for a patch $P$ is computed by summing $\mathbf{D}_{\text{total}}$ over all spatial locations $(h,w)$ and all three color channels $c$ within that patch: $\text{IE}(P) = \sum_{c=1}^{3} \sum_{(h,w) \in P} \mathbf{D}_{\text{total}}^{(c, h, w)}$. A lower IE value indicates a smoother, less textured region. This operation is implemented as a summation over the designated tensor dimensions (`dim=[1, 2, 3]`).

This design allows the IE metric to capture fine-grained, channel-wise texture anomalies that are critical for distinguishing subtle generative artifacts.

\section{Additional Ablation Studies and Results}
\subsection{Comprehensive Analysis of Patch Selection}
\begin{table}[ht]
    \centering
    \caption{Performance of Different Patch Selection Strategies on DIFF-Intra. Only the best achieved accuracy (\textbf{Best Acc(\%)}) is retained for brevity.}
    \label{table:patch_performance_simplified}
    \begin{tabular}{ccc|c}
        \toprule
        \textbf{Type of Patches} & \textbf{Number of Patches} & \textbf{Dataset Size} & \textbf{Acc(\%)} \\
        \midrule
        max & 16 & all & 94.91 \\
        max & 16 & 20\% & 89.94 \\
        max & 64 & all & 94.26 \\
        max & 64 & 20\% & 83.78 \\
        \midrule
        mid & 4 & all & 98.66 \\
        mid & 4 & 20\% & 89.56 \\
        mid & 16 & all & 97.93 \\
        mid & 16 & 20\% & 86.99 \\
        mid & 64 & all & 96.94 \\
        mid & 64 & 20\% & 86.11 \\
        \midrule
        min & 1 & all & 98.13 \\
        min & 4 & all & 97.86 \\
        min & 4 & 20\% & 90.17 \\
        \rowcolor{gray!20} min & 16 & all & 98.85 \\
        \rowcolor{gray!20} min & 16 & 20\% & 97.17 \\
        min & 64 & all & 97.67 \\
        min & 256 & all & 92.65 \\
        \midrule
        ori & - & all & 98.66 \\
        ori & - & 20\% & 90.86 \\
        random & 16 & all & 97.62 \\
        \bottomrule
    \end{tabular}
\end{table}

To validate our patch selection strategy, we experiment with different criteria based on \textbf{Image Entropy}, including patches with minimum, maximum, and central entropy, as well as the entire image and randomly selected patches. All tests are conducted using ResNet50 with varying numbers of patches and dataset sizes.

As shown in \cref{table:patch_performance_simplified}, our findings reveal the following:

1. \textbf{Patch-based detection is highly effective}: Even random patch selection achieves competitive accuracy (97.62\%), confirming that localized regions contain sufficient forensic cues to distinguish AI-generated content.

2. \textbf{Minimum-entropy patches consistently outperform others}: The best configuration—16-patch division with minimum-entropy selection—yields a peak accuracy of \textbf{98.85\%}, and maintains high performance even with only 20\% of the training data (97.17\%).

3. \textbf{High-entropy patches are suboptimal}: Despite their complexity, maximum-entropy regions likely introduce irrelevant noise, reducing generalization.

These results support the design rationale of EGPL: low-entropy regions are more robust and artifact-prone, making them ideal candidates for patch-based detection. While random or center patches also perform well, our entropy-guided approach offers greater consistency and data efficiency.

\subsection{Sensitivity Analysis of the VIB Weight $\beta$}
\label{supp:beta_ablation}
The variational information bottleneck (VIB) module is regulated by the hyperparameter $\beta$, which controls the trade-off between preserving task-relevant information and compressing the latent representation. While $\beta = 10^{-3}$ is a commonly adopted default, we conduct a focused sensitivity analysis to verify its optimality for our specific task.

\textbf{Experimental Setup:} We trained the full EIB-Net (with EGPL selecting the minimum entropy patch from a $16$-patch split) on the DIFF-Intra subset. All other hyperparameters (learning rate, batch size, data augmentation) were held constant. We evaluated three values: $\beta \in \{10^{-4}, 10^{-3}, 10^{-2}\}$.

\textbf{Results and Analysis:} The validation accuracy for each $\beta$ is reported in \cref{tab:beta_sensitivity}. The results confirm that $\beta = 10^{-3}$ yields the best performance ($97.17\%$). A weaker regularization ($\beta=10^{-4}$) leads to under-compression and lower accuracy ($96.75\%$), while a stronger regularization ($\beta=10^{-2}$) likely causes excessive information loss, also harming performance ($96.86\%$). This analysis justifies our selection of $\beta = 10^{-3}$ for all main experiments.

\begin{table}[ht]
    \centering
    \caption{Sensitivity analysis of the VIB regularization parameter $\beta$ on the DIFF-Intra subset.}
    \label{tab:beta_sensitivity}
    \begin{tabular}{@{}ccc@{}}
        \toprule
        $\beta$ & \textbf{Validation Accuracy} (\%) & \textbf{Relative Performance} \\
        \midrule
        $1 \times 10^{-4}$ & 96.86 & $-$0.31\% \\
        $1 \times 10^{-3} (Ours)$ & \textbf{97.17} & -- \\
            $1 \times 10^{-2}$ & 96.75 & $-$0.42\% \\
        \bottomrule
    \end{tabular}
\end{table}

\subsection{Ablation on Classification Loss Components}
\label{supp:loss_ablation}
To decouple the contributions of different components within our chosen Jensen-Shannon Divergence (JSD) loss, we conduct a progressive ablation study on the DIFF-Intra subset. The JSD loss can be decomposed into a Label Smoothing Cross-Entropy (LSCE) base and an additional JSD consistency regularization term. We compare three settings:
\begin{itemize}
    \item \textbf{CE}: Standard Cross-Entropy loss.
    \item \textbf{LSCE only}: Only the Label Smoothing component ($\epsilon=0.1$).
    \item \textbf{JSD (Ours)}: The full loss combining LSCE and JSD regularization ($\alpha=12$).
\end{itemize}

\begin{table}[h]
    \centering
    \caption{Progressive ablation on the classification loss. The full JSD loss yields the best performance.}
    \label{tab:loss_ablation}
    \begin{tabular}{@{}lcc@{}}
        \toprule
        \textbf{Loss Function} & \textbf{Validation Accuracy (\%)} & \textbf{Relative Performance}\\
        \midrule
        CE & 95.60 & -- \\
        LSCE only  & 96.25 & $+$0.65\% \\
        JSD (Ours) & \textbf{97.17} & $+$1.57\%\ \\
        \bottomrule
    \end{tabular}
\end{table}

\noindent\textbf{Analysis:} The results demonstrate that both components contribute to the final performance. Label smoothing provides a stable gain of $+0.65\%$ over the CE baseline by mitigating over-confidence. On top of that, the JSD consistency regularization further improves accuracy by $+0.88\%$, leading to a total gain of $+1.57\%$. This confirms that the full JSD loss is not merely effective due to label smoothing, but also because of its inherent self-consistency regularization, which aligns well with our goal of learning robust and generalizable features.


\endgroup


\begin{thebibliography}{10}

\bibitem{GAN}
Ian Goodfellow, Jean Pouget-Abadie, Mehdi Mirza, Bing Xu, David Warde-Farley, Sherjil Ozair, Aaron Courville, and Yoshua Bengio,
\newblock ``Generative adversarial networks,''
\newblock {\em Communications of the ACM}, vol. 63, no. 11, pp. 139--144, 2020.

\bibitem{SDv1}
Robin Rombach, Andreas Blattmann, Dominik Lorenz, Patrick Esser, and Bj{\"o}rn Ommer,
\newblock ``High-resolution image synthesis with latent diffusion models,''
\newblock in {\em CVPR}, 2022, pp. 10684--10695.

\bibitem{Midjourney}
Midjourney,
\newblock ``Midjourney,'' \url{https://www.midjourney.com/home/}, 2022.

\bibitem{ff++}
Andreas Rossler, Davide Cozzolino, Luisa Verdoliva, Christian Riess, Justus Thies, and Matthias Nie{\ss}ner,
\newblock ``Faceforensics++: Learning to detect manipulated facial images,''
\newblock in {\em ICCV}, 2019, pp. 1--11.

\bibitem{kohler1970gestalt}
Wolfgang K{\"o}hler,
\newblock {\em Gestalt psychology: An introduction to new concepts in modern psychology}, vol.~18,
\newblock WW Norton \& Company, 1970.

\bibitem{multi_task}
Huy~H Nguyen, Fuming Fang, Junichi Yamagishi, and Isao Echizen,
\newblock ``Multi-task learning for detecting and segmenting manipulated facial images and videos,''
\newblock in {\em BTAS}, 2019, pp. 1--8.

\bibitem{cnn3}
Run Wang, Felix Juefei-Xu, Lei Ma, Xiaofei Xie, Yihao Huang, Jian Wang, and Yang Liu,
\newblock ``Fakespotter: a simple yet robust baseline for spotting ai-synthesized fake faces,''
\newblock in {\em IJCAI}, 2021, pp. 3444--3451.

\bibitem{lisiam}
Jian Wang, Yunlian Sun, and Jinhui Tang,
\newblock ``Lisiam: Localization invariance siamese network for deepfake detection,''
\newblock {\em IEEE TIFS}, vol. 17, pp. 2425--2436, 2022.

\bibitem{duan2025dual}
Junxian Duan, Siyu Liu, Yiming Hao, Huaibo Huang, and Ran He,
\newblock ``Dual frequency-guided spatiotemporal feature learning for face forgery detection,''
\newblock {\em IEEE Transactions on Biometrics, Behavior, and Identity Science}, 2025.

\bibitem{DIRE}
Zhendong Wang, Jianmin Bao, Wengang Zhou, Weilun Wang, Hezhen Hu, Hong Chen, and Houqiang Li,
\newblock ``Dire for diffusion-generated image detection,''
\newblock in {\em ICCV}, 2023, pp. 22445--22455.

\bibitem{AEROBLADE}
Jonas Ricker, Denis Lukovnikov, and Asja Fischer,
\newblock ``Aeroblade: Training-free detection of latent diffusion images using autoencoder reconstruction error,''
\newblock in {\em CVPR}, 2024, pp. 9130--9140.

\bibitem{LATENTTRACER}
Zhenting Wang, Vikash Sehwag, Chen Chen, Lingjuan Lyu, Dimitris~N Metaxas, and Shiqing Ma,
\newblock ``How to trace latent generative model generated images without artificial watermark?,''
\newblock in {\em ICML}, 2024.

\bibitem{NPR}
Chuangchuang Tan, Yao Zhao, Shikui Wei, Guanghua Gu, Ping Liu, and Yunchao Wei,
\newblock ``Rethinking the up-sampling operations in cnn-based generative network for generalizable deepfake detection,''
\newblock in {\em CVPR}, 2024, pp. 28130--28139.

\bibitem{zhu2023gendet}
Mingjian Zhu, Hanting Chen, Mouxiao Huang, Wei Li, Hailin Hu, Jie Hu, and Yunhe Wang,
\newblock ``Gendet: Towards good generalizations for ai-generated image detection,''
\newblock {\em arXiv preprint arXiv:2312.08880}, 2023.

\bibitem{Shadow}
Ayush Sarkar, Hanlin Mai, Amitabh Mahapatra, Svetlana Lazebnik, David~A Forsyth, and Anand Bhattad,
\newblock ``Shadows don't lie and lines can't bend! generative models don't know projective geometry... for now,''
\newblock in {\em CVPR}, 2024, pp. 28140--28149.

\bibitem{tishby2000information}
Naftali Tishby, Fernando~C Pereira, and William Bialek,
\newblock ``The information bottleneck method,''
\newblock {\em arXiv preprint physics/0004057}, 2000.

\bibitem{alemi2016deep}
Alexander~A Alemi, Ian Fischer, Joshua~V Dillon, and Kevin Murphy,
\newblock ``Deep variational information bottleneck,''
\newblock {\em arXiv preprint arXiv:1612.00410}, 2016.

\bibitem{achille2018information}
Alessandro Achille and Stefano Soatto,
\newblock ``Information dropout: Learning optimal representations through noisy computation,''
\newblock {\em TPAMI}, vol. 40, no. 12, pp. 2897--2905, 2018.

\bibitem{higgins2017beta}
Irina Higgins, Loic Matthey, Arka Pal, Christopher Burgess, Xavier Glorot, Matthew Botvinick, Shakir Mohamed, and Alexander Lerchner,
\newblock ``beta-vae: Learning basic visual concepts with a constrained variational framework,''
\newblock in {\em ICLR}, 2017.

\bibitem{versteeg2019bottleneck}
Greg Ver~Steeg and Aram Galstyan,
\newblock ``The bottleneck principle: Enhancing interpretability in deep neural networks,''
\newblock in {\em AAAI}, 2019, vol.~33, pp. 3656--3663.

\bibitem{sun2021ibood}
Xinwei Sun, Wenhu Chen, and William~Yang Wang,
\newblock ``Information bottleneck learning for out-of-distribution generalization,''
\newblock in {\em ICML}. PMLR, 2021, pp. 9940--9950.

\bibitem{wang2021ibadv}
Binghui Wang and Neil~Zhenqiang Gong,
\newblock ``Information bottleneck approach to adversarial robustness,''
\newblock in {\em ICLR}, 2021.

\bibitem{solopova2025}
Veronika Solopova, Lucas Schmidt, and Dorothea Kolossa,
\newblock ``Extending information bottleneck attribution to video sequences,'' 2025.

\bibitem{DIFF}
Harry Cheng, Yangyang Guo, Tianyi Wang, Liqiang Nie, and Mohan Kankanhalli,
\newblock ``Diffusion facial forgery detection,''
\newblock {\em arXiv preprint arXiv:2401.15859}, 2024.

\bibitem{GenImage}
Mingjian Zhu, Hanting Chen, Qiangyu Yan, Xudong Huang, Guanyu Lin, Wei Li, Zhijun Tu, Hailin Hu, Jie Hu, and Yunhe Wang,
\newblock ``Genimage: A million-scale benchmark for detecting ai-generated image,''
\newblock {\em NIPS}, vol. 36, 2024.

\end{thebibliography}

\begin{thebibliography}{10}

\bibitem{supp:DIRE}
Zhendong Wang, Jianmin Bao, Wengang Zhou, Weilun Wang, Hezhen Hu, Hong Chen, and Houqiang Li,
\newblock ``Dire for diffusion-generated image detection,''
\newblock in {\em ICCV}, 2023, pp. 22445--22455.

\bibitem{supp:LSUN}
Fisher Yu, Ari Seff, Yinda Zhang, Shuran Song, Thomas Funkhouser, and Jianxiong Xiao,
\newblock ``Lsun: Construction of a large-scale image dataset using deep learning with humans in the loop,''
\newblock {\em arXiv preprint arXiv:1506.03365}, 2015.

\bibitem{supp:celebHQ}
Tero Karras, Timo Aila, Samuli Laine, and Jaakko Lehtinen,
\newblock ``Progressive growing of gans for improved quality, stability, and variation,''
\newblock in {\em ICLR}, 2018.

\bibitem{supp:GenImage}
Mingjian Zhu, Hanting Chen, Qiangyu Yan, Xudong Huang, Guanyu Lin, Wei Li, Zhijun Tu, Hailin Hu, Jie Hu, and Yunhe Wang,
\newblock ``Genimage: A million-scale benchmark for detecting ai-generated image,''
\newblock {\em NIPS}, vol. 36, 2024.

\bibitem{supp:nichol2022glide}
Alexander~Quinn Nichol, Prafulla Dhariwal, Aditya Ramesh, Pranav Shyam, Pamela Mishkin, Bob Mcgrew, Ilya Sutskever, and Mark Chen,
\newblock ``Glide: Towards photorealistic image generation and editing with text-guided diffusion models,''
\newblock in {\em ICML}. PMLR, 2022, pp. 16784--16804.

\bibitem{supp:Wukong}
Wukong,
\newblock ``Wukong,'' \url{https://xihe.mindspore.cn/modelzoo/wukong}, 2022.

\bibitem{supp:Midjourney}
Midjourney,
\newblock ``Midjourney,'' \url{https://www.midjourney.com/home/}, 2022.

\bibitem{supp:LDMs}
Robin Rombach, Andreas Blattmann, Dominik Lorenz, Patrick Esser, and Bj{\"o}rn Ommer,
\newblock ``High-resolution image synthesis with latent diffusion models,''
\newblock in {\em CVPR}, 2022, pp. 10684--10695.

\bibitem{supp:VQDM}
Shuyang Gu, Dong Chen, Jianmin Bao, Fang Wen, Bo~Zhang, Dongdong Chen, Lu~Yuan, and Baining Guo,
\newblock ``Vector quantized diffusion model for text-to-image synthesis,''
\newblock in {\em CVPR}, 2022, pp. 10696--10706.

\bibitem{supp:classifierbootstrap}
Prafulla Dhariwal and Alexander Nichol,
\newblock ``Diffusion models beat gans on image synthesis,''
\newblock {\em NIPS}, vol. 34, pp. 8780--8794, 2021.

\bibitem{supp:DIFF}
Harry Cheng, Yangyang Guo, Tianyi Wang, Liqiang Nie, and Mohan Kankanhalli,
\newblock ``Diffusion facial forgery detection,''
\newblock {\em arXiv preprint arXiv:2401.15859}, 2024.

\bibitem{supp:podellsdxl}
Dustin Podell, Zion English, Kyle Lacey, Andreas Blattmann, Tim Dockhorn, Jonas M{\"u}ller, Joe Penna, and Robin Rombach,
\newblock ``Sdxl: Improving latent diffusion models for high-resolution image synthesis,''
\newblock in {\em ICLR}, 2024.

\bibitem{supp:freedom}
Jiwen Yu, Yinhuai Wang, Chen Zhao, Bernard Ghanem, and Jian Zhang,
\newblock ``Freedom: Training-free energy-guided conditional diffusion model,''
\newblock in {\em ICCV}, 2023, pp. 23174--23184.

\bibitem{supp:HPS}
Xiaoshi Wu, Keqiang Sun, Feng Zhu, Rui Zhao, and Hongsheng Li,
\newblock ``Better aligning text-to-image models with human preference,''
\newblock {\em arXiv preprint arXiv:2303.14420}, vol. 1, no. 3, 2023.

\bibitem{supp:hu2021lora}
Edward~J Hu, Yelong Shen, Phillip Wallis, Zeyuan Allen-Zhu, Yuanzhi Li, Shean Wang, Lu~Wang, and Weizhu Chen,
\newblock ``Lora: Low-rank adaptation of large language models,''
\newblock {\em arXiv preprint arXiv:2106.09685}, 2021.

\bibitem{supp:ruiz2023dreambooth}
Nataniel Ruiz, Yuanzhen Li, Varun Jampani, Yael Pritch, Michael Rubinstein, and Kfir Aberman,
\newblock ``Dreambooth: Fine tuning text-to-image diffusion models for subject-driven generation,''
\newblock in {\em CVPR}, 2023, pp. 22500--22510.

\bibitem{supp:kim2022diffface}
Kihong Kim, Yunho Kim, Seokju Cho, Junyoung Seo, Jisu Nam, Kychul Lee, Seungryong Kim, and KwangHee Lee,
\newblock ``Diffface: Diffusion-based face swapping with facial guidance,''
\newblock {\em arXiv e-prints}, pp. arXiv--2212, 2022.

\bibitem{supp:kim2023dcface}
Minchul Kim, Feng Liu, Anil Jain, and Xiaoming Liu,
\newblock ``Dcface: Synthetic face generation with dual condition diffusion model,''
\newblock in {\em CVPR}, 2023, pp. 12715--12725.

\bibitem{supp:kawar2023imagic}
Bahjat Kawar, Shiran Zada, Oran Lang, Omer Tov, Huiwen Chang, Tali Dekel, Inbar Mosseri, and Michal Irani,
\newblock ``Imagic: Text-based real image editing with diffusion models,''
\newblock in {\em CVPR}, 2023, pp. 6007--6017.

\bibitem{supp:huang2023collaborative}
Ziqi Huang, Kelvin~CK Chan, Yuming Jiang, and Ziwei Liu,
\newblock ``Collaborative diffusion for multi-modal face generation and editing,''
\newblock in {\em CVPR}, 2023, pp. 6080--6090.

\bibitem{supp:wu2023latent}
Chen~Henry Wu and Fernando De~la Torre,
\newblock ``A latent space of stochastic diffusion models for zero-shot image editing and guidance,''
\newblock in {\em ICCV}, 2023, pp. 7378--7387.

\bibitem{supp:DIFF_real1}
Yeqi Bai, Tao Ma, Lipo Wang, and Zhenjie Zhang,
\newblock ``Speech fusion to face: Bridging the gap between human's vocal characteristics and facial imaging,''
\newblock in {\em ACM MM}, 2022, pp. 2042--2050.

\bibitem{supp:DIFF_real2}
Joon~Son Chung, Arsha Nagrani, and Andrew Zisserman,
\newblock ``Voxceleb2: Deep speaker recognition,''
\newblock {\em arXiv preprint arXiv:1806.05622}, 2018.

\bibitem{supp:DIFF_real3}
Cheng-Han Lee, Ziwei Liu, Lingyun Wu, and Ping Luo,
\newblock ``Maskgan: Towards diverse and interactive facial image manipulation,''
\newblock in {\em CVPR}, 2020, pp. 5549--5558.

\bibitem{supp:resnet}
Kaiming He, Xiangyu Zhang, Shaoqing Ren, and Jian Sun,
\newblock ``Deep residual learning for image recognition,''
\newblock in {\em CVPR}, 2016, pp. 770--778.

\bibitem{supp:rmt}
Qihang Fan, Huaibo Huang, Mingrui Chen, Hongmin Liu, and Ran He,
\newblock ``Rmt: Retentive networks meet vision transformers,''
\newblock in {\em CVPR}, 2024, pp. 5641--5651.

\bibitem{supp:vit}
Alexey Dosovitskiy, Lucas Beyer, Alexander Kolesnikov, Dirk Weissenborn, Xiaohua Zhai, Thomas Unterthiner, Mostafa Dehghani, Matthias Minderer, Georg Heigold, Sylvain Gelly, Jakob Uszkoreit, and Neil Houlsby,
\newblock ``An image is worth 16x16 words: Transformers for image recognition at scale,''
\newblock {\em ICLR}, 2021.

\bibitem{supp:xception}
Fran{\c{c}}ois Chollet,
\newblock ``Xception: Deep learning with depthwise separable convolutions,''
\newblock in {\em CVPR}, 2017, pp. 1251--1258.

\end{thebibliography}
\end{document}